\documentclass{article}

\usepackage{microtype}
\usepackage{graphicx}
\usepackage{subfigure}
\usepackage{booktabs} 

\usepackage{hyperref}

\usepackage[accepted]{icml2019}

\usepackage{url}
\usepackage{amsmath}
\usepackage{amsthm}
\usepackage{graphicx}
\usepackage{tabularx}
\usepackage{multirow}
\usepackage{booktabs}
\usepackage{amsfonts}
\usepackage{color}
\usepackage{float}
\usepackage{epstopdf}
\usepackage{threeparttable}
\usepackage[english]{babel}

\newtheorem{definition}{\textbf{Definition}}

\newtheorem{proposition}{\textbf{Proposition}}

\usepackage{array}
\newcommand{\PreserveBackslash}[1]{\let\temp=\\#1\let\\=\temp}
\newcolumntype{C}[1]{>{\PreserveBackslash\centering}p{#1}}
\newcolumntype{R}[1]{>{\PreserveBackslash\raggedleft}p{#1}}
\newcolumntype{L}[1]{>{\PreserveBackslash\raggedright}p{#1}}

\begin{document}

\twocolumn[
\icmltitle{Differentiable Architecture Search with Ensemble Gumbel-Softmax}



\begin{icmlauthorlist}
\icmlauthor{Jianlong Chang}{ed1,ed2}
\icmlauthor{Xinbang Zhang}{ed1,ed2}
\icmlauthor{Yiwen Guo}{intel}
\icmlauthor{Gaofeng Meng}{ed1}
\icmlauthor{Shiming Xiang}{ed1,ed2}
\icmlauthor{Chunhong Pan}{ed1}
\end{icmlauthorlist}

\icmlaffiliation{ed1}{National Laboratory of Pattern Recognition, Institute of Automation, Chinese Academy of Sciences}
\icmlaffiliation{ed2}{School of Artificial Intelligence, University of Chinese Academy of Sciences}
\icmlaffiliation{intel}{Intel Labs China}

\icmlcorrespondingauthor{Jianlong Chang}{jianlong.chang@nlpr.ia.ac.cn}

\icmlkeywords{Machine Learning, ICML}

\vskip 0.3in
]



\printAffiliationsAndNotice{}  

\begin{abstract}
For network architecture search (NAS), it is crucial but challenging to simultaneously guarantee both effectiveness and efficiency. Towards achieving this goal, we develop a differentiable NAS solution, where the search space includes arbitrary feed-forward  network consisting of the predefined number of connections. Benefiting from a proposed \textit{ensemble Gumbel-Softmax} estimator, our method optimizes both the architecture of a deep network and its parameters in the same round of backward propagation, yielding an end-to-end mechanism of searching network architectures. Extensive experiments on a variety of popular datasets strongly evidence that our method is capable of discovering high-performance architectures, while guaranteeing the requisite efficiency during searching. 
\end{abstract}

\section{Introduction}

In the era of deep learning, how to design reasonable network architecture for specific problems is a challenging task. Designing architecture with state-of-the-art performance typically requires substantial efforts from human experts. In order to eliminate such exhausting engineering, much research has been devoted to automatically searching architectures, namely neural architecture search (NAS), which has achieved significant successes in a multitude of fields, including image classification~\cite{DBLP:conf/cvpr/ZophVSL18, DBLP:journals/corr/abs-1802-01548, DBLP:conf/eccv/LiuZNSHLFYHM18, DBLP:journals/corr/abs-1806-09055, DBLP:journals/corr/abs-1812-09926}, semantic segmentation~\cite{DBLP:conf/nips/ChenCZPZSAS18} and object detection~\cite{DBLP:conf/cvpr/ZophVSL18}.

So far, there exist three basic frameworks that have gained a growing interest, \textit{i.e.}, evolution-based NAS, reinforcement learning-based NAS, and gradient-based NAS. 
Primitively, evolution-based NAS employs evolution algorithms to jointly learn architectures and parameters in networks, such as NEAT~\cite{stanley2002evolving}. Due to the uneconomical searching strategy in evolution algorithms, evolution-based NAS inherently requires tremendous time and resource consumption. For instance, it takes 3150 GPU days for AmoebaNet~\cite{DBLP:journals/corr/abs-1802-01548} to achieve the state-of-the-art performance in comparison of human-designed architectures. To boost the process, reinforcement learning-based NAS using back-propagation seems to be a natural and plausible choice. When compared with some evolution-based NAS, many reinforcement learning-based NAS methods like ENAS~\cite{DBLP:conf/icml/PhamGZLD18} can dramatically reduce the time and resource consumption despite of the similar optimization mechanism. However, since the reinforcement learning-based NAS is generally established as a Markov decision process, temporal-difference is utilized to take structural decisions. As a result, the reward in NAS will only be observable until the architecture is chosen and the network is tested for accuracy, which is always subject to delayed rewards in temporal-difference learning~\cite{DBLP:journals/corr/abs-1806-07857}. To eliminate such deficiency, gradient-based NAS methods including DARTS~\cite{DBLP:journals/corr/abs-1806-09055} and SNAS~\cite{DBLP:journals/corr/abs-1812-09926} are recently presented. They try to convert the concrete search space into a continuous one, so that architectures and parameters can be well-optimized by gradient descent.

In this work, we develop an effective and efficient NAS method, Differentiable ARchiTecture Search with Ensemble Gumbel-Softmax (DARTS-EGS), which is capable of discovering more diversified network architectures, while maintaining the differentiability of a promising NAS pipeline. In order to guarantee the diversity of network architectures along the search path, we represent the whole search space with binary codes. 
Benefiting from such artful modeling, the whole feed-forward networks consisting of the predefined number of ordered nodes are included into the search space in our model. To maintain the prerequisite efficiency in searching, we develop ensemble Gumbel-Softmax to replace the traditional softmax for yielding a differentiable end-to-end mechanism. That is, our ensemble Gumbel-Softmax is in a position to perform any structural decisions like policy gradient in the reinforcement-learning, but more efficient than temporal difference learning considering delayed rewards.

To sum up, the main contributions of this work are:
\begin{itemize}
  \item By generalizing the traditional Gumbel-Softmax, we develop an ensemble Gumbel-Softmax, which provides a successful attempt to effectively and efficiently perform any structural decisions like policy gradient in the reinforcement-learning, with higher efficiency.
  \item Benefiting from the ensemble Gumbel-Softmax, the search space can be dramatically increased while maintaining the requisite efficiency in searching, which yields an end-to-end  mechanism to identify the network architecture with requisite computational complexity.
  \item Extensive experiments verify that our model outperforms current models in searching high-performance convolutional architectures for image classification and recurrent architectures for language modeling.
\end{itemize}


\section{Related Work}

Recently, discovering neural architecture automatically has raised great interest in both academia and industry~\cite{DBLP:journals/corr/BelloPLNB16, baker2017accelerating, DBLP:journals/corr/abs-1708-05344, DBLP:journals/corr/abs-1708-05038, DBLP:conf/ijcai/SuganumaSN18, DBLP:conf/eccv/VeitB18}. Nowadays, neural architecture search method can be roughly divided into three class according to searching method~\cite{DBLP:journals/corr/abs-1808-05377}, \textit{i.e.}, evolution-based NAS, reinforcement learning-based NAS, and gradient-based NAS.

Evolution-based neural architecture search methods~\cite{DBLP:conf/icml/RealMSSSTLK17, DBLP:journals/corr/MiikkulainenLMR17, DBLP:journals/corr/abs-1802-01548, elsken2018efficient, DBLP:journals/corr/abs-1803-06744} utilize evolution algorithms to generate neural architecture, automatically. In~\cite{DBLP:conf/icml/RealMSSSTLK17,DBLP:journals/corr/abs-1802-01548,elsken2018efficient}, a large CNN architecture space is explored, and modifications like inserting layer, adjusting filter size and adding identity mapping are designed as mutations in evolution. Despite the remarkable achievements, their methods require huge computation resource and less practical in large scale.

Reinforcement learning based neural architecture search methods are prevalent in recent work~\cite{DBLP:journals/corr/ZophL16, DBLP:conf/icml/BelloZVL17, DBLP:conf/cvpr/ZophVSL18, DBLP:conf/cvpr/ZhongYWSL18,DBLP:conf/icml/PhamGZLD18}. In the pioneering work~\cite{DBLP:journals/corr/ZophL16}, an RNN network is utilized as the controller to decide the type and parameters of layers sequentially. The controller is trained by reinforcement learning with the accuracy of the generated architecture designed as reward. Although it achieves impressive results, the searching process is computational hunger and 800 GPUs are required. Based on~\cite{DBLP:journals/corr/ZophL16}, several methods have been proposed to accelerate the search process. Specifically, \cite{DBLP:conf/cvpr/ZophVSL18,DBLP:conf/cvpr/ZhongYWSL18} diminish the search space by searching the architecture of block and then stack the searched block to generate final network. In~\cite{DBLP:conf/icml/PhamGZLD18}, the weights of network are shared among child models, saving searching time by reducing the cost of getting and evaluation. Additionally, a series of well-performance methods have also been explored, including the progressive search~\cite{DBLP:conf/eccv/LiuZNSHLFYHM18} and multi-objective optimization~\cite{DBLP:journals/corr/abs-1807-11626,DBLP:journals/corr/abs-1806-10332}.

Contrary to treating architecture search as black-box optimization problem, gradient based neural architecture search methods utilized the gradient obtained in the training process to optimize neural architecture~\cite{shin2018differentiable, DBLP:conf/nips/LuoTQCL18, DBLP:journals/corr/abs-1806-09055, DBLP:journals/corr/abs-1812-09926}. Typically, NAO~\cite{DBLP:conf/nips/LuoTQCL18} utilizes RNN networks as the encoder and decoder to map architectures into a continuous network embeddings space and conduct optimization in this space with gradient-based method. Another typical method DARTS~\cite{DBLP:journals/corr/abs-1806-09055} chooses the best connection between nodes from a candidate primitive set by employing a softmax classifier. Although DARTS achieves impressive results, the discrete process is not totally differentiable and the connection between two nodes are limited to a single one in the candidate primitive set merely.

\section{Methodology}


In the following, Section~\ref{Search Space} introduces our notations, defines the search space and presents our strategy of encoding network architectures with binary codes. Section~\ref{Continuous Relaxation} introduces a conceptually intuitive yet powerful relaxation for searching in the discrete search space. Section~\ref{Optimization with Ensemble Gumbel-Softmax} elaborates the proposed ensemble Gumbel-Softmax estimator, which plays a key role in constructing our scheme for jointly optimizing the architecture and its weights. Finally, Section~\ref{Parameter Leaning and Architecture Sampling} details the objective in our model.

\subsection{Search Space}\label{Search Space}

To balance the optimality and efficiency in NAS, we search for computation cells that constitute the whole network, following prior works~\cite{DBLP:conf/cvpr/ZophVSL18, DBLP:conf/eccv/LiuZNSHLFYHM18, DBLP:journals/corr/abs-1802-01548, DBLP:conf/icml/PhamGZLD18, DBLP:journals/corr/abs-1806-09055, DBLP:journals/corr/abs-1812-09926, DBLP:journals/corr/abs-1812-00332}. In essense, every such cell is a sub-network, which can be naturally considered as a directed acyclic graph (DAG) consisting of an ordered sequence of nodes. To cover abundant network architectures, our search space is set as the whole space of DAGs with the predefined number of nodes.

Without loss of generality and for simplicity, we denote a cell with $n$ nodes as $\mathcal{C}=\{e^{(i,j)}|1\leq i<j\leq n\}$, where $e^{(i,j)}$ indicates a directed edge from the $i$-th node to the $j$-th node. Corresponding to each directed edge $e^{(i,j)}$, there are a set of candidate primitive operations $\mathcal{O}=\{o_{1},\cdots,o_{K}\}$, such as convolution, pooling, identity, and zero\footnote{``Zero" means no connection between two nodes.}, as defined in DARTS~\cite{DBLP:journals/corr/abs-1806-09055}. With these operations, the output at the $j$-th node can be formulated as
\begin{equation}
\begin{aligned}\label{basic_formula1}
x^{(j)}=\sum\limits_{i<j}o^{(i,j)}(x^{(i)})
\end{aligned}
\end{equation}
where $x^{(i)}$ denotes the input from the $i$-th node, and $o^{(i,j)}(\cdot)$ is a function applied to $x^{(i)}$ which can be decomposed into a superposition of primitive operations in $\mathcal{O}$, \textit{i.e.},
\begin{equation}
\begin{aligned}\label{basic_formula2}
&o^{(i,j)}(x^{(i)})=\sum\limits_{k=1}^{K}\alpha_{k}^{(i,j)}\cdot o_{k}(x^{(i)})\\
&s.t.~~~\alpha_{k}^{(i,j)}\in\{0,1\}
\end{aligned}
\end{equation}
where $o_{k}(\cdot)$ is the $k$-th candidate primitive operation in $\mathcal{O}$, and $\alpha_{k}^{(i,j)}$ signifies a binary and thus discrete weight to indicate whether the operation $o_{k}(\cdot)$ is utilized on the edge $e^{(i,j)}$. For clarity, we introduce the binary set $\mathcal{A}=\{\alpha_{k}^{(i,j)}|1\leq i<j\leq n,1\leq k\leq K\}$ as the network architecture code to represent a cell and $\mathcal{A}^{(i,j)}=\{\alpha_{k}^{(i,j)}|1\leq k\leq K\}$ as the edge architecture code to represent the structure of the edge $e^{(i,j)}$. To reveal the capability of above encoding for representing cells, we have the following proposition
\begin{proposition}\label{proposition1}
For an arbitrary feed-forward network consisting of limited numbers of ordered nodes and candidate operations in $\mathcal{O}$, there is one and only one architecture code $\mathcal{A}$ that corresponds to it.
\end{proposition}

Proposition~\ref{proposition1} guarantees the uniqueness of our model encoding. In the meanwhile, it also implies that our search space includes the whole set of feed-forward networks and guarantees the expressive capability of the formulation in Eq.~(\ref{basic_formula2}). With such modeling, many previously defined seach spaces can be regarded as subsets of ours. For example, our modeling shall degrade into that of DARTS~\cite{DBLP:journals/corr/abs-1806-09055} and SNAS~\cite{DBLP:journals/corr/abs-1812-09926} when $\sum_{k=1}^{K}\alpha_{k}^{(i,j)}=1$ is further introduced as a constraint. Limited by this constraint, only one operation will be chosen on each edge, which can be considered as a one-category problem.

\subsection{Relaxation of The Search Space}\label{Continuous Relaxation}

Benefiting from the uniqueness property of our architecture code $\mathcal{A}$, as verified in Proposition~\ref{proposition1}, the task of learning the cell can therefore be converted to approaching the optimal code. Yet, as mentioned, one major obstacle to approaching efficient NAS is the difficulty of optimization in the discrete space. In this subsection, we shall introduce our strategy for searching in the continuous space as a proxy. Without loss of generality, we elaborate the strategy on the edge $e^{(i,j)}$.

To make the search space continuous, we relax the categorical choice of a set of particular operations as follows
\begin{equation}
\begin{aligned}\label{relaxation}
&\tilde{o}^{(i,j)}(x^{(i)})=\sum\limits_{k=1}^{K} f \left(p_{k}^{(i,j)}\right)\cdot o_{k}(x^{(i)}),\\
&s.t.~~~\sum\nolimits_{k=1}^{K}p_{k}^{(i,j)}=1,\\
&~~~~~~~~~p_{k}^{(i,j)}\geq0,~\forall1\leq k\leq K,\\
&~~~~~~~~~ f \left(p_{k}^{(i,j)}\right)\in\{0,1\},~\forall1\leq k\leq K,
\end{aligned}
\end{equation}
where $p_{k}^{(i,j)}$ denotes the probability of choosing the $k$-th operation on the edge $e^{(i,j)}$, and $ f (\cdot)$ represents a binary function that suffices to map a probability vector to a binary code and pass gradients in a continuous manner. We will discuss more about it in Section~\ref{Optimization with Ensemble Gumbel-Softmax} and explain how we manage to pass gradients with it. Specifically, $ f (\cdot)$ is chosen to be a monotonic increasing function in our method, \textit{i.e.},
\begin{equation}
 f (p_{k_{1}}^{(i,j)})\leq f (p_{k_{2}}^{(i,j)}),\quad \textrm{if}\ p_{k_{1}}^{(i,j)}\leq p_{k_{2}}^{(i,j)}.
\end{equation}
By substituting $\alpha_{k}^{(i,j)}$ with $ f (p_{k}^{(i,j)})$ and considering $p_{k}^{(i,j)}$ instead as the variable to be optimized, we have successfully achieved a continuous relaxation. Benefiting from the flexibility of our formulation, it is capable of modeling multi-category problems and incorporating resource constraint seamlessly, as will be discussed.

\begin{figure*}[!ht]
\begin{center}
\centerline{\includegraphics[width=14cm]{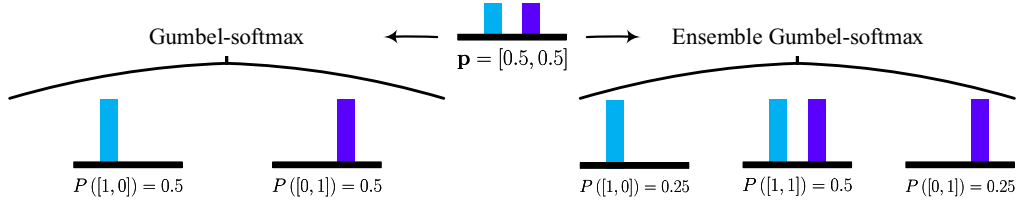}}
\vskip -0.25in
\caption{A visualized comparison between Gumbel-Softmax (left) and ensemble Gumbel-Softmax (right, $ M =2$). For a probability vector $\mathbf{p}=[0.5,0.5]$, Gumbel-Softmax solely pertains to sample only two binary codes with the same probability, \textit{i.e.}, $P([1,0])=P([0,1])=0.5$. In contrast, our ensemble Gumbel-Softmax is capable of sampling more diversified binary codes, \textit{i.e.}, $[1,0]$, $[1,1]$ and $[0,1]$. Furthermore, the probabilities of sampling these binary codes are logical. Typically, it is conceptually intuitive that the probability of sampling $[1,1]$ is larger than the probabilities of sampling the others since the probabilities in $\mathbf{p}=[0.5,0.5]$ are equal to each other.}
\vskip -0.35in
\label{egs}
\end{center}
\end{figure*}

\subsubsection{Modeling Multi-Category Problems}

Different from the traditional probability vector that solely pertains to manage one-category problem, the formulation in Eq.~(\ref{relaxation}) is in a position to solve multi-category problems. This superiority inherently comes from the monotonicity of the function $ f (\cdot)$. It endows the formulation with a capability of modeling more general relationships between different categories, not limited to the incompatibility purely. 

\subsubsection{Modeling Resource Constraint}

In NAS, desired methods should not only show effectiveness (\textit{i.e.}, test-set accuracy) but also possess superior efficiency (\textit{i.e.}, search cost). 
Aiming at a high performing NAS framework, we integrate the two core parts together and represent $p_{k}^{(i,j)}$ as
\begin{equation}
\begin{aligned}\label{resource_constraint}
p_{k}^{(i,j)}=\lambda\cdot h_{k}^{(i,j)}+(1-\lambda)\cdot l_{k}^{(i,j)},
\end{aligned}
\end{equation}
where $h_{k}^{(i,j)}$ and $l_{k}^{(i,j)}$ represent the credits in terms of effectiveness and efficiency when choosing the $k$-th operation on the edge $e^{(i,j)}$, and $0\leq\lambda\leq1$ is a hyper-parameter for balancing the two parts. In practice, the validation-set accuracy and search time are employed to represent the effectiveness and efficiency of our model respectively, and $\lambda=0.5$ is always fixed in this work.

\subsection{Optimization with Ensemble Gumbel-Softmax}\label{Optimization with Ensemble Gumbel-Softmax}

Although the relaxation presented in Section~\ref{Continuous Relaxation} makes the search space continuous, how to define the binary function $ f (\cdot)$ as desired to map each of the probabilities to a binary code needs to be sorted out. In principle, it is still a problem of learing to take concrete decisions, which is unfortunately indifferentiable and has no gradients almost everywhere. To leverage the gradient information as in generic deep learning, we introduce an ensemble Gumbel-Softmax estimator to optimize the problem with a principled approximation. As such, the back-propagation algorithm can be directly adopted in a end-to-end manner, yielding an efficient and effective searching mechanism.

\begin{figure*}[!ht]
\begin{center}
\centerline{\includegraphics[width=14cm]{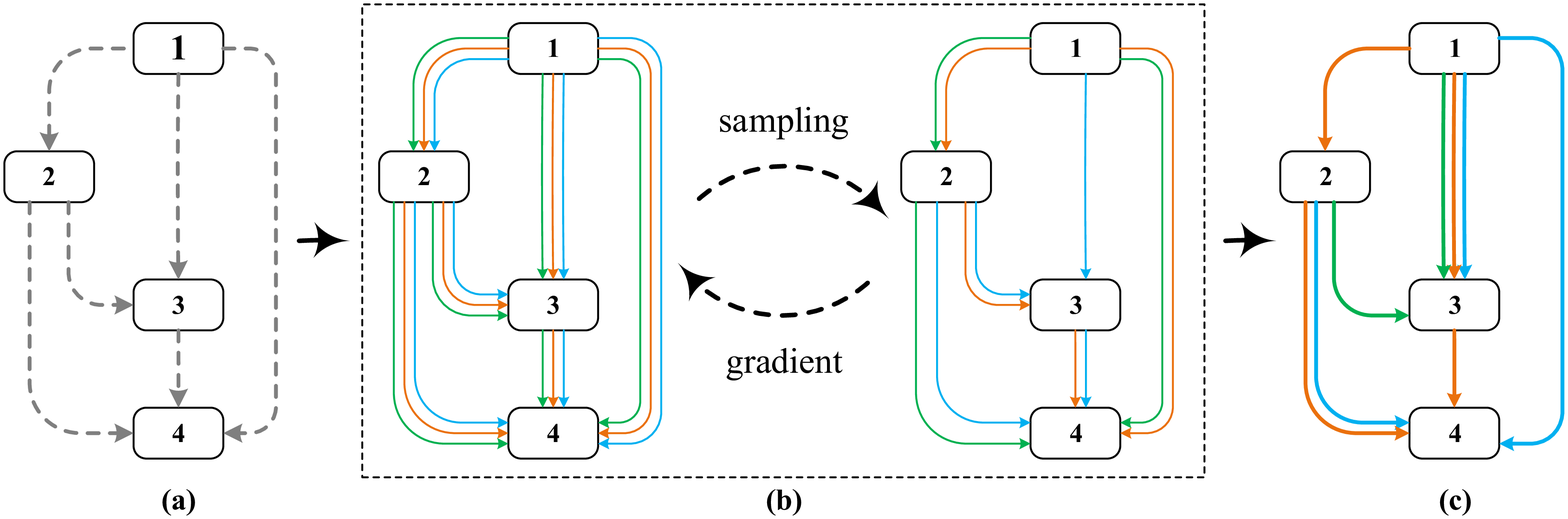}}
\vskip -0.3in
\caption{A conceptual visualization for the searching process within DARTS-EGS. (a) First, a cell (\textit{i.e.}, directed acyclic graph) consisting of four ordered nodes is predefined. (b) During the forward propagation, with three candidate primitive operations (\textit{i.e.}, green, orange and cyan lines), our ensemble Gumbel-Softmax is employed to sample a network in a differentiable manner. During the backward propagation, the standard back-propagation algorithm is utilized to simultaneously calculate the gradients of the both architecture and network. (c) Finally, the details of the cell can be sampled with our ensemble Gumbel-Softmax and utilized to handle specific tasks.}
\vskip -0.3in
\label{main}
\end{center}
\end{figure*}

\subsubsection{Gumbel-Softmax}\label{Gumbel-Softmax}

A natural formulation for representing discrete variable is to use the categorical distribution. However, partially due to the inability to back-propagate information through samples, it seems rarely applied in deep learning. In this work, we resort to the Gumbel-Max trick~\cite{gumbel1954statistical} for enabling back-propagation and and representing the process of taking decision as sampling from a categorical distribution, in order to perform NAS in a principled way. Specifically, given a probability vector $\mathbf{p}=[p_{1},\cdots,p_{K}]$ and a discrete random variable with $P(L=k)\propto p_{k}$, we sample from the discrete variable $L$ by introducing the Gumbel random variables. To be more specific, we let
\begin{equation}
\begin{aligned}\label{gumbelmax}
L=\arg\max\limits_{k\in\{1,\cdots,K\}}\log p_{k}+G_{k},  
\end{aligned}
\end{equation}
where $\{G_{k}\}_{k\leq K}$ is a sequence of the standard Gumbel random variables, and they are typically sampled from the Gumbel distribution $G=-\log(-\log(X))$ with $X\sim U[0,1]$. An obstacle to directly using such approach in our problem is that the argmax operation is not really continuous. One straightforward way of dealing with this problem is to replace the argmax operation with a softmax~\cite{DBLP:journals/corr/JangGP16,DBLP:journals/corr/MaddisonMT16}. Formally, the Gumbel-Softmax (GS) estimation can be expressed as
\begin{equation}
\begin{aligned}\label{gumbelsoftmax}
\hat{L}_{k}=\frac{\exp\left(\left(\log p_{k}+G_{k}\right)/\tau\right)}{\sum_{k=1}^{K}\exp\left(\left(\log p_{k}+G_{k}\right)/\tau\right)},
\end{aligned}
\end{equation}
where $\hat{L}_{k}$ indicates the probability that $p_{k}$ is the maximal entry in $\mathbf{p}$, and $\tau$ is a temperature. When $\tau\rightarrow0$, $[\hat{L}_{1},\cdots,\hat{L}_{K}]$ converges to an one-hot vector, and in the other extreme it will become a discrete uniform distribution with $\tau\rightarrow+\infty$.

From the expression in Eq.~(\ref{gumbelsoftmax}), we see that the traditional Gumbel-Softmax pertains solely to deal with the problems that only one category requires to be determined. In NAS, however, an optimal architecture may require multiple operations on an edge, considering the practical significance~\cite{DBLP:conf/cvpr/HeZRS16,DBLP:conf/cvpr/SzegedyVISW16}. For instance, the residual module $\mathbf{y}=F(\mathbf{x})+I(\mathbf{x})$ in ResNets~\cite{DBLP:conf/cvpr/HeZRS16} consists of two operations with a learnable mapping $F(\cdot)$ and the identity $I(\cdot)$. That is, choosing different operations in $\mathcal{O}$ may not be mutually exclusive but compatible. One direct way of handling this limitation is to map all possible operation combinations to $2^{K}$-dimensional vectors, where $K$ is the number of candidate operations in $\mathcal{O}$. However, it seems difficult to search architectures efficiently when there are many candidate operations (\textit{i.e.}, $K$ is large).

\subsubsection{Ensemble Gumbel-Softmax}\label{Ensemble Gumbel-Softmax}

In order to address the aforementioned limitation in the traditional Gumbel-Softmax, we propose our Ensemble Gumbel-Softmax (EGS) estimator. With the assistance of this estimator, we will be able to choose multiple operations on each edge by sampling more diversified codes, not limited to the one-hot codes merely. To this end, we start by recoding binary codes. 
For the code $\mathcal{A}^{(i,j)}=\{\alpha_{k}^{(i,j)}|1\leq k\leq K\}$ on an edge $e^{(i,j)}$, the whole architecture information included in $\mathcal{A}^{(i,j)}$ can be recoded into a superposition of $K$ one-hot vectors, \textit{i.e.},
\begin{equation}
\begin{aligned}\label{decomposion}
\mathcal{A}^{(i,j)}\simeq\sum\limits_{k=1}^{K} \mathbf{p}_{k}^{(i,j)}
\end{aligned}
\end{equation}
where $\mathbf{p}_{k}^{(i,j)}\in\mathbb{R}^{K}$ is a $K$-dimensional one-hot vector that uniquely corresponds to the operation $o_{k}\in\mathcal{O}$. Such equivalence relationship reveals that compositing the results sampled from Gumbel-Softmax may be a possible way to sample $\mathcal{A}^{(i,j)}$ straightforward, although Gumbel-Softmax is capable of sampling $\mathbf{p}_{k}^{(i,j)}$, as described in Section~\ref{Gumbel-Softmax}.

For an effective and efficient sampler, we composite the whole sampling results for every $\mathbf{p}_{o}^{(i,j)}$. That is, an ensemble of multiple Gumbel-Softmax samplers is profound for sampling diversified binary codes, \textit{i.e.},
\begin{definition}\label{definition}
For a $K$-dimensional probability vector $\mathbf{p}=[p_{1},\cdots,p_{K}]$ and $ M $ one-hot vectors $\{\mathbf{z}^{(1)},\cdots,\mathbf{z}^{( M )}\}$ sampled from $\mathbf{p}$, the $K$-dimensional binary code $\tilde{\mathbf{p}}$ sampled with ensemble Gumbel-Softmax is
\begin{equation*}
\begin{aligned}
\tilde{p}_{k}=\max\limits_{1\leq i\leq M }(z_{k}^{(i)}),1\leq k\leq K
\end{aligned}
\end{equation*}
where $M$ is the number of sampling times, $\tilde{p}_{k}$ is the $k$-th element in $\tilde{\mathbf{p}}$, and $z_{k}^{(i)}$ indicates the $k$-th element in $\mathbf{z}^{(i)}$.
\end{definition}

\begin{figure*}[!ht]
\begin{center}
\centerline{\includegraphics[width=17cm]{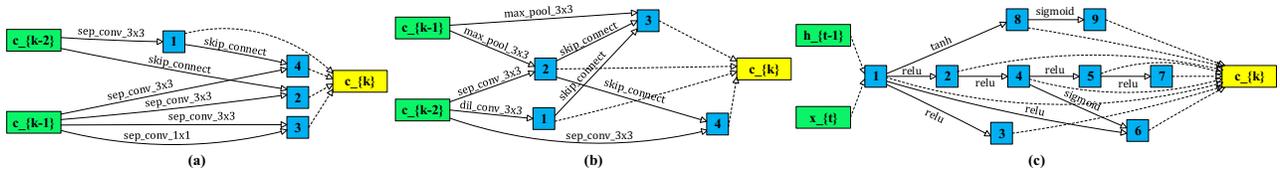}}
\vskip -0.25in
\caption{Three cells learned by DARTS-EGS ($M=1$). (a) Normal cell learned on CIFAR-10. (b) Reduction cell learned on CIFAR-10. (c) Recurrent cell learned on PTB. Specifically, the green and yellow nodes indicate the input and output, respectively. The solid and dashed lines means the learnable and predefined operations, respectively.}
\vskip -0.3in
\label{result_arch}
\end{center}
\end{figure*}

\begin{table*}[t]\footnotesize
\renewcommand{\arraystretch}{0.95}
\caption{Comparison with state-of-the-art image classifiers on CIFAR-10 (lower error rate is better).}
\label{cifar10}
\centering
\begin{threeparttable}
\begin{tabular}{@{}L{6cm} C{1.5cm} C{1.5cm} C{1.5cm} C{1.5cm} C{2.5cm}}
\toprule
\multirow{2}*{Architecture} & Test Error & Params & Search Cost & \multirow{2}*{Ops} & \multirow{2}*{Search}\\
\cmidrule(lr){2-2}\cmidrule(lr){3-3}\cmidrule(lr){4-4}
 & (\%) & (M) & (GPU days) & \\
\midrule
DenseNet-BC~\cite{DBLP:conf/cvpr/HuangLMW17} & 3.46 & 25.6 & - & - & manual\\
\midrule
NASNet-A + cutout~\cite{DBLP:conf/cvpr/ZophVSL18} & 2.65 & 3.3 & 2000 & 13 & RL\\
BlockQNN~\cite{DBLP:conf/cvpr/ZhongYWSL18} & 3.54 & 39.8 & 96 & 8 & RL\\
ENAS + cutout~\cite{DBLP:conf/icml/PhamGZLD18} & 2.89 & 4.6 & 0.5 & 6 & RL\\
AmoebaNet-A~\cite{DBLP:journals/corr/abs-1802-01548} & 3.34 & 3.2 & 3150 & 19 & evolution\\
AmoebaNet-B + cutout~\cite{DBLP:journals/corr/abs-1802-01548} & 2.55 & 2.8 & 3150 & 19 & evolution\\
Hierarchical evolution~\cite{DBLP:journals/corr/abs-1711-00436} & 3.75 & 15.7 & 300 & 6  & evolution\\
PNAS~\cite{DBLP:conf/eccv/LiuZNSHLFYHM18} & 3.41 & 3.2 & 225 & 8 & SMBO\\
DARTS ($1$-th order) + cutout~\cite{DBLP:journals/corr/abs-1806-09055} & 3.00 & 3.3 & 1.5 & 7 & gradient-based\\
DARTS ($2$-th order) + cutout~\cite{DBLP:journals/corr/abs-1806-09055} & 2.76 & 3.3 & 4 & 7 & gradient-based\\
SNAS + mild + cutout~\cite{DBLP:journals/corr/abs-1812-09926} & 2.98 & 2.9 & 1.5 & - & gradient-based\\
SNAS + moderate + cutout~\cite{DBLP:journals/corr/abs-1812-09926} & 2.85 & 2.8 & 1.5 & - & gradient-based\\
SNAS + aggressive + cutout~\cite{DBLP:journals/corr/abs-1812-09926} & 3.10 & 2.3 & 1.5 & - & gradient-based\\
\midrule
Random search baseline + cutout & 3.29 & 3.2 & 4 & 7 & random\\
DARTS-EGS ($M=4$) & 3.01 & 2.6 & 1 & 7 & gradient-based\\
DARTS-EGS ($M=7$) & 2.79 & 2.9 & 1 & 7 & gradient-based\\
\bottomrule
\end{tabular}
\end{threeparttable}
\vspace{-0.3cm}
\end{table*}

\subsubsection{Understanding Our EGS}\label{Understanding Ensemble GS}

To reveal the serviceability and sampling capability of our ensemble Gumbel-Softmax, from the definition, two basic propositions are given in the following.

First, whether an element in the sampled binary code is one depends on the probability at the corresponding location.
\begin{proposition}\label{proposition2}
For arbitrary probability vector $\mathbf{p}=[p_{1},\cdots,p_{K}]$ and number of sampling times $M$, we have
\begin{equation*}
\begin{aligned}
P(\tilde{p}_{k}=1)\propto p_{k}
\end{aligned}
\end{equation*}
where $P(\tilde{p}_{k}=1)$ is the probability of $\tilde{p}_{k}=1$. Furthermore,
\begin{equation*}
\begin{aligned}
P(\tilde{p}_{k1}=1)\leq P(\tilde{p}_{k2}=1) \Leftrightarrow p_{k1}\leq p_{k2}
\end{aligned}
\end{equation*}
where $P(\tilde{p}_{k1}=1)=P(\tilde{p}_{k2}=1)\Leftrightarrow p_{k1}=p_{k2}$.
\end{proposition}
Proposition~\ref{proposition2} implies that our ensemble Gumbel-Softmax is a mono-tonic increasing function in terms of probability, and in a position to act as the function $ f (\cdot)$ in Eq.~(\ref{relaxation}).

Second, the capability of sampling binary codes is determined on $ M $, \textit{i.e.}, the number of one-hot vectors sampled form Gumbel-Softmax. Specifically, we have
\begin{proposition}\label{proposition3}
For arbitrary probability vector $\mathbf{p}=[p_{1},\cdots,p_{K}]$ and number of sampling times $M$, the ensemble Gumbel-Softmax is capable of sampling $\binom{K}{ M }\times (2^{ M }-1)$ different binary codes, which includes the whole binary codes with up to $ M $ ones and at least $1$ one.
\end{proposition}
Proposition~\ref{proposition3} indicates that the sampling capability of increases exponentially with $ M $. In practice, larger $ M $ is always employed to deal with more complex tasks for effect, and smaller one can be utilized to search more lightweight networks for efficiency.

Synthetically, our ensemble Gumbel-Softmax is not only available for searching network architectures but also sampling capability for guaranteeing the performance in theory. In Figure~\ref{egs}, a visualized comparison between Gumbel-Softmax and ensemble Gumbel-Softmax intuitively shows that our ensemble Gumbel-Softmax is more excellent than the traditional Gumbel-Softmax, in terms of both sampling capability and rationality in practice.

\subsection{Parameter Leaning and Architecture Sampling}\label{Parameter Leaning and Architecture Sampling}

To eliminate the compromise of the NAS pipeline, the parameters and architectures will be simultaneously optimized interms of ensemble Gumbel-Softmax. Analogous to architecture search in~\cite{DBLP:journals/corr/ZophL16, DBLP:conf/cvpr/ZophVSL18, DBLP:conf/icml/PhamGZLD18, DBLP:journals/corr/abs-1711-00436, DBLP:journals/corr/abs-1802-01548, DBLP:journals/corr/abs-1806-09055}, the validation set performance is considered as the reward in our model, but using an end-to-end differentiable manner. Denote the training loss as $\mathcal{L}_{w}(\mathcal{A})$. The goal in architecture search is to find a high-performance architecture, \textit{i.e.},
\begin{equation}
\begin{aligned}
\min\limits_{w,\alpha} \mathbb{E}_{\mathcal{A}\sim p_{\alpha}(\mathcal{A})}[\mathcal{L}_{w}(\mathcal{A})]
\end{aligned}.
\end{equation}

The main process of optimizing this objective is to minimize the expected performance of architectures sampled with $p_{\alpha}(\mathcal{A})$. That is, the network $\mathcal{A}$ is first sampled with $p_{\alpha}(\mathcal{A})$. Afterward, the loss on the training dataset can be calculated by forward propagation. Relying on this loss, the gradients of the network architecture parameter $\alpha$ and the network parameter $w$ are yielded to modify these parameters better. Because of the differentiability, our model can be trained end-to-end by the standard back-propagation algorithm. In the end, the network architecture $\mathcal{A}$ is identified by sampling with $p_{\alpha}(\mathcal{A})$, and the network parameter $w$ is estimated by retraining on the training set. A conceptual visualization of our DARTS-EGS model is illustrated in Figure~\ref{main}.

\section{Experiments}

In this section, we systematically carry out extensive experiments to verify the capability of our model in discovering high-performance convolutional networks for image classification and recurrent networks for language modeling. For each task, the experiments consist of two stages, following with the previous work~\cite{DBLP:journals/corr/abs-1806-09055, DBLP:journals/corr/abs-1812-09926}. First, we search for the cell architectures based on our ensemble Gumbel-Softmax and find the best cells according to their validation performance. Second, the transferability of the best cells learned on CIFAR-10~\cite{krizhevsky2009learning} and Penn Tree Bank (PTB)~\cite{taylor2003penn} are investigated by employing them on large datasets, \textit{i.e.}, classification on ImageNet~\cite{DBLP:conf/cvpr/DengDSLL009} and language modeling on WikiText-2 (WT2)~\cite{DBLP:journals/corr/MerityXBS16}, respectively. As a greatly improved work of DARTS, specifically, the experimental settings always inherit from it, except some special settings in each experiment.

\subsection{Image Classification}

\begin{table*}[t]\footnotesize
\renewcommand{\arraystretch}{0.95}
\caption{Comparison with state-of-the-art image classifiers on ImageNet in the mobile setting (lower error rate is better).}
\label{imagenet}
\centering
\begin{threeparttable}
\begin{tabular}{@{}L{6.5cm} C{1.25cm} C{1.25cm} C{1.5cm} C{1.5cm} C{2.5cm}}
\toprule
\multirow{2}*{Architecture} & \multicolumn{2}{c}{Test Error (\%)} & Params & Search Cost & \multirow{2}*{Search}\\
\cmidrule(lr){2-3}\cmidrule(lr){4-4}\cmidrule(lr){5-5}
& Top 1 & Top 5 & (M) & (GPU days) & \\
\midrule
Inception-v1~\cite{DBLP:conf/cvpr/SzegedyLJSRAEVR15} & 30.2 & 10.1 & 6.6 & - & manual\\
MobileNet~\cite{DBLP:journals/corr/HowardZCKWWAA17} & 29.4 & 10.5 & 4.2 & - & manual\\
ShuffleNet-v1 2$\times$~\cite{DBLP:conf/cvpr/ZhangZLS18} & 26.3 & - & $\sim$5 & - & manual\\
ShuffleNet-v2 2$\times$~\cite{DBLP:conf/eccv/MaZZS18} & 25.1 & - & $\sim$5 & - & manual\\
\midrule
AmoebaNet-A~\cite{DBLP:journals/corr/abs-1802-01548} & 25.5 & 8.0 & 5.1 & 3150 & evolution\\
AmoebaNet-B~\cite{DBLP:journals/corr/abs-1802-01548} & 26.0 & 8.5 & 5.3 & 3150 & evolution\\
AmoebaNet-C~\cite{DBLP:journals/corr/abs-1802-01548} & 24.3 & 7.6 & 6.4 & 3150 & evolution\\
PNAS~\cite{DBLP:conf/eccv/LiuZNSHLFYHM18} & 25.8 & 8.1 & 5.1 & $\sim$225 & SMBO\\
DARTS (searched on CIFAR-10)~\cite{DBLP:journals/corr/abs-1806-09055} & 26.7 & 8.7 & 4.7 & 4 & gradient-based\\
SNAS (mild constraint)~\cite{DBLP:journals/corr/abs-1812-09926} & 27.3 & 9.2 & 4.3 & 1.5 & gradient-based\\
NASNet-A~\cite{DBLP:conf/cvpr/ZophVSL18} & 26.0 & 8.4 & 5.3 & 2000 & RL\\
NASNet-B~\cite{DBLP:conf/cvpr/ZophVSL18} & 27.2 & 8.7 & 5.3 & 2000 & RL\\
NASNet-C~\cite{DBLP:conf/cvpr/ZophVSL18} & 27.5 & 9.0 & 4.9 & 2000 & RL\\
\midrule
DARTS-EGS ($M=4$) & 25.7 & 8.5 & 4.3 & 1.5 & gradient-based\\
DARTS-EGS ($M=7$) & 24.9 & 8.1 & 4.7 & 1.5 & gradient-based\\
\bottomrule
\end{tabular}
\end{threeparttable}
\vspace{-0.25cm}
\end{table*}

\subsubsection{Architecture Search on CIFAR-10}

In our experiments, eight typical operations are included in the candidate primitive set $\mathcal{O}$: $3\times3$ and $5\times5$ separable convolutions, $3\times3$ and $5\times5$ dilated separable convolutions, $3\times3$ max pooling, $3\times3$ average pooling, identity, and zero. In order to preserve their spatial resolution, all operations are of stride one, and the convolutional feature maps are padded if necessary. In ensemble Gumbel-Softmax, the number of sampling times $M=7$ is set for a rich search space. Following the settings in the previous work~\cite{DBLP:conf/eccv/LiuZNSHLFYHM18, DBLP:journals/corr/abs-1806-09055, DBLP:journals/corr/abs-1812-09926}, the ReLU-Conv-BN order is utilized in the whole convolution operations, and every separable convolution is always applied twice.

The settings of nodes in our convolutional cell are also following the previous work~\cite{DBLP:conf/cvpr/ZophVSL18, DBLP:journals/corr/abs-1802-01548, DBLP:conf/eccv/LiuZNSHLFYHM18, DBLP:journals/corr/abs-1806-09055}. Specifically, every cell consists of $n=7$ nodes, among which the output node is defined as the depthwise concatenation of all the intermediate nodes. More larger networks are always established by stacking multiple cells together. In the $k$-th cell, the first and second nodes are set equally to the outputs in the $(k-2)$-th and $(k-1)$-th cells respectively, with $1\times1$ convolution as necessary. Furthermore, the reduction cell with the architecture coding $\mathcal{A}_{reduce}$ is utilized at the $1/3$ and $2/3$ of the total depth of the network. The rest of cells are the normal cell with the architecture coding $\mathcal{A}_{normal}$.

\subsubsection{Architecture Evaluation on CIFAR-10}\label{Architecture Evaluation on CIFAR-10}

To evaluate the selected architecture, a large network of 20 cells is trained from scratch for 600 epoches with batch size 96 and report its performance on the test set. We add additional enhancements include cutout with size 16, path dropout of probability 0.2 and auxiliary towers with weight 0.4 following exiting works for fair comparison. We report the mean of 4 independent runs for our full model.

Figure~\ref{result_arch} and Table~\ref{cifar10} give the searched architectures and classification results on CIFAR-10, which shows that our DARTS-EGS achieces comparable results with the state-of-the-art with less computation resources. Such a good performance verifies that DARTS-EGS can effectively and efficiently search worthy architectures for classification. In DARTS-EGS, furthermore, higher accuracy is yielded when $M=7$ compared with $M=4$. This scenario is in accordance with our motivation that more richer search spaces is beneficial for searching architectures.

\begin{table*}[t]\footnotesize
\renewcommand{\arraystretch}{0.875}
\caption{Comparison with state-of-the-art language models on PTB (lower perplexity is better).}
\label{ptb}
\centering
\begin{threeparttable}
\begin{tabular}{@{}L{6cm} C{0.8cm} C{0.8cm} C{0.8cm} C{2.4cm} C{0.8cm} C{2.5cm}}
\toprule
\multirow{2}*{Architecture} & \multicolumn{2}{c}{Perplexity} & Params & Search Cost & \multirow{2}*{Ops} & \multirow{2}*{Search}\\
\cmidrule(lr){2-3}\cmidrule(lr){4-4}\cmidrule(lr){5-5}
& valid & test & (M) & (GPU days) & \\
\midrule
Variational RHN~\cite{DBLP:journals/corr/ZillySKS16} & 67.9 & 65.4 & 23 & - & - & manual\\
LSTM~\cite{DBLP:journals/corr/abs-1708-02182} & 60.7 & 58.8 & 24 & - & - & manual\\
LSTM + skip connections~\cite{DBLP:journals/corr/MelisDB17} & 60.9 & 58.3 & 24 & - & - & manual\\
LSTM + 15 softmax experts~\cite{DBLP:journals/corr/abs-1711-03953} & 58.1 & 56.0 & 22 & - & - &manual\\
DARTS (first order)~\cite{DBLP:journals/corr/abs-1806-09055} & 60.2 & 57.6 & 23 & 0.5 & 4 & gradient-based\\
DARTS (second order)~\cite{DBLP:journals/corr/abs-1806-09055} & 58.1 & 55.7 & 23 & 1 & 4 & gradient-based\\
NAS~\cite{DBLP:journals/corr/ZophL16} & - & 64.0 & 25 & 1e4 CPU days & 4 & RL\\
ENAS~\cite{DBLP:conf/icml/PhamGZLD18} & 68.3 & 63.1 & 24 & 0.5 & 4 & RL\\
\midrule
Random search baseline & 61.8 & 59.4 & 23 & 2 & 4 & random\\
DARTS-EGS ($M=4$) & 58.3 & 56.2 & 22 & 0.5 & 4 & gradient-based\\
DARTS-EGS ($M=7$) & 57.1 & 55.3 & 23 & 0.5 & 4 & gradient-based\\
\bottomrule
\end{tabular}
\end{threeparttable}
\vspace{-0.5cm}
\end{table*}

\begin{table*}[t]\footnotesize
\renewcommand{\arraystretch}{0.9}
\caption{Comparison with state-of-the-art language models on WT2 (lower perplexity rate is better).}
\label{wt2}
\centering
\begin{threeparttable}
\begin{tabular}{@{}L{6cm} C{1.5cm} C{1.5cm} C{1.5cm} C{1.5cm} C{2.5cm}}
\toprule
\multirow{2}*{Architecture} & \multicolumn{2}{c}{Perplexity} & Params & Search Cost & \multirow{2}*{Search}\\
\cmidrule(lr){2-3}\cmidrule(lr){4-4}\cmidrule(lr){5-5}
& valid & test & (M) & (GPU days) & \\
\midrule
LSTM + augmented loss~\cite{DBLP:journals/corr/InanKS16} & 91.5 & 87.0 & 28 & - & manual\\
LSTM + cache pointer~\cite{DBLP:journals/corr/GraveJU16} & - & 68.9 & - & - & manual\\
LSTM~\cite{DBLP:journals/corr/abs-1708-02182} & 69.1 & 66.0 & 33 & - & manual\\
LSTM + skip connections~\cite{DBLP:journals/corr/MelisDB17} & 69.1 & 65.9 & 24 & - & manual\\
LSTM + 15 softmax experts~\cite{DBLP:journals/corr/abs-1711-03953} & 66.0 & 63.3 & 33 & - & manual\\
DARTS  (searched on PTB)~\cite{DBLP:journals/corr/abs-1806-09055}  & 69.5 & 66.9  & 33 & 1 & gradient-based\\
ENAS (searched on PTB)~\cite{DBLP:conf/icml/PhamGZLD18} & 72.4 & 70.4 & 33 & 0.5 & RL\\
\midrule
DARTS-EGS ($M=4$) & 67.3 & 64.6  & 33 & 1 & gradient-based\\
DARTS-EGS ($M=7$) & 66.5 & 64.2  & 33 & 1 & gradient-based\\
\bottomrule
\end{tabular}
\end{threeparttable}
\vspace{-0.25cm}
\end{table*}

\begin{table*}[t]\footnotesize
\renewcommand{\arraystretch}{0.9}
\caption{Sensitivity to number of sampling times on CIFAR-10 (lower error rate is better).}
\label{sensitivity_number}
\centering
\begin{threeparttable}
\begin{tabular}{@{}L{5cm} C{0.8cm} C{0.8cm} C{0.8cm} C{0.8cm} C{0.8cm} C{0.8cm} C{0.8cm} C{0.8cm} C{0.8cm}}
\toprule
Number of Sampling Times ($M$) & 1 & 2 & 3 & 4 & 5 & 6 & 7 & 8 & 9 \\
\midrule
Test Error (\%) & 3.38 & 3.22 & 3.09 & 3.05 & 3.01 & 2.84 & 2.79 & 2.74 & 2.73 \\
Params (M) & 2.29 & 2.51 & 2.56 & 2.60 & 2.72 & 2.84 & 2.79 & 2.90 & 3.02 \\
\bottomrule
\end{tabular}
\end{threeparttable}
\vspace{-0.25cm}
\end{table*}

\subsubsection{Transferability Evaluation on ImageNet}\label{Transferability Evaluation on ImageNet}

We apply mobile setting where the input image size is 224$\times$224 and the number of multiply-add operations of the model is restricted to be under 600M. A network of 14 cells is trained for 250 epoches with batch size 128, weight decay 3$\times10^{-5}$ and poly learning rate scheduler with initial learning rate 0.1. Label smoothing~\cite{DBLP:conf/cvpr/SzegedyVISW16} and auxiliary loss~\cite{DBLP:conf/aistats/LeeXGZT15} are used during training. Other hyperparameters follow~\cite{DBLP:journals/corr/abs-1806-09055}.

In Table~\ref{imagenet}, we report the quantitative results on ImageNet. Note that the cell searched on CIFAR-10 can be smoothly employed to deal with the large-scale classification task. Compared with other gradient-based NAS methods, furthermore, greater margins are yielded on ImageNet. A possible reason is that more complex architectures can be searched in DARTS-EGS because of the larger search space. Consequently, such more complex architectures handle more complex task on ImageNet better.

\subsection{Language Modeling}

\subsubsection{Architecture Search on PTB}

In the language modeling task, our model is employed to search suitable activation function between nodes. Following the setting in~\cite{DBLP:conf/cvpr/ZophVSL18, DBLP:conf/icml/PhamGZLD18, DBLP:journals/corr/abs-1806-09055}, five popular functions are considered in the candidate primitive set $\mathcal{O}$, such as sigmoid, tanh, relu, identity, and zero. In addition, there are $n=12$ nodes in the recurrent cell, and the number of sampling times $M=4$ is set in ensemble Gumbel-Softmax for a rich search space. Similar to ENAS~\cite{DBLP:conf/icml/PhamGZLD18} and DARTS~\cite{DBLP:journals/corr/abs-1806-09055}, in cells, the very first intermediate node is obtained by linearly transforming the two input nodes, adding up the results and then passing through the tanh function, and the rest of activation functions are learned with our model and enhanced with the highway bypass~\cite{DBLP:conf/icml/ZillySKS17}. The batch normalization~\cite{DBLP:conf/icml/IoffeS15} in each node to prevent gradient explosion during architecture search, and disable it during architecture evaluation. Our recurrent network consists of only a single cell, \textit{i.e.}, we do not assume any repetitive patterns within the recurrent architecture.

\subsubsection{Architecture Evaluation on PTB}

We train a single-layer recurrent network with searched cell for 1600 epoches with batch size 64 using averaged SGD, Both the embedding and the hidden sizes are set to 850 to ensure our model size is comparable with other baselines. Other hyper-parameters are set following~\cite{DBLP:journals/corr/abs-1806-09055}. Note that the model is not fine-tuned at the end of the optimization, nor do we use any additional enhancements for a fair comparison.

Table~\ref{ptb} lists the results in this experiment. From the table, we observe that DARTS-EGS also is in a position to search recurrent architectures effectively. It empirically shows that the back-propagation algorithm can guide DARTS-EGS to hit a preferable architecture in a larger search space, while maintaining the requisite efficiency. Similar to the conclusion in Section~\ref{Architecture Evaluation on CIFAR-10}, lower perplexity is achieved when the larger $M$ is employed, which verifies that more richer search spaces is also valuable for recurrent architectures .

\subsubsection{Transferability Evaluation on WT2}\label{Transferability Evaluation on WT2}

Different from the setting on PTB, we apply embdding hidden sizes 700, weight decay 5$\times10^{-7}$, and hidden-node variational dropout 0.15. Other hyperparameters remain the same in the experiment on PTB. In Table~\ref{wt2}, the results on WT2 indicates that the transferability is also retentive on recurrent architectures. Conclusively, the consistent results in Section~\ref{Transferability Evaluation on ImageNet} and~\ref{Transferability Evaluation on WT2} strongly guarantee the transferability on both convolutional and recurrent architectures.

\subsection{Ablation study}

\subsubsection{Sensitivity to Number of Sampling Times}

We perform experiments on CIFAR-10 to analyze the sensitivities to the number of sampling time $M$. Table~\ref{sensitivity_number} gives the results in this experiment. From this table, it can be observed that larger $M$ indicates higher performance, while more parameters will be introduced as $M$ increases. This is in accordance with the statement in Proposition~\ref{proposition3}, \textit{i.e.}, more capable networks might be found with larger M to get higher performance.

\subsubsection{Performance on Semantic Segmentation}

We also validate the capability of our method on a more complex task, \textit{i.e.}, semantic segmentation on VOC-2012. Compared with DARTS ($75.4\%$), DARTS-EGS achieves better performance ($76.8\%$) with the larger margin $+1.4\%$ than the classification on ImageNet. Such results demonstrate that DARTS-EGS may have more prominent superiority on more complex tasks, not just toy tasks.

\section{Conclusion}

We present a powerful framework to search network architectures, DARTS-EGS, which is capable of covering more diversified network architectures, while maintaining the differentiability of the NAS pipeline. For this purpose, the network architectures are represented with arbitrary discrete binary codes, guaranteeing the reach of search space. In order to ensure the efficiency in searching, ensemble Gumbel-Softmax is developed to search architectures in a differentiable end-to-end manner. By searching with the standard back-propagation, DARTS-EGS is able to outperform the state-of-the-art architecture search methods on various tasks, with remarkable efficiency.

Future work may include search the whole networks with our ensemble Gumbel-Softmax and injecting our ensemble Gumbel-Softmax into deep models to handle other machine learning tasks. For the first work, the sampling capability of ensemble Gumbel-Softmax guarantees the practicability of searching any networks, but how to improve the efficiency remains to be solved. For the second work, the differentiability of ensemble Gumbel-Softmax indicates that it can be utilized anywhere in networks. Relying on such insight, an interesting direction is to recast the clustering process into our ensemble Gumbel-Softmax. By aggregating inputs in each cluster, conclusively, a general pooling for both deep networks and deep graph networks~\cite{DBLP:journals/corr/abs-1806-01261,Chang_2018_NIPS} can be developed to deal with Euclidean and non-Euclidean structured data uniformly.

\bibliography{darts-egs}
\bibliographystyle{icml2019}
\end{document}